# CFPNet: Channel-wise Feature Pyramid for Real-Time Semantic Segmentation

Ange Lou, Murray Loew
Department of Biomedical Engineering, The George Washington University, Washington DC, USA

## Abstract

*Real-time semantic segmentation is playing a more important role in computer vision, due to the growing demand for mobile devices and autonomous driving. Therefore, it is very important to achieve a good trade-off among performance, model size and inference speed. In this paper, we propose a Channel-wise Feature Pyramid (CFP) module to balance those factors. Based on the CFP module, we built CFPNet for real-time semantic segmentation which applied a series of dilated convolution channels to extract effective features. Experiments on Cityscapes and CamVid datasets show that the proposed CFPNet achieves an effective combination of those factors. For the Cityscapes test dataset, CFPNet achievse 70.1% class-wise mIoU with only 0.55 million parameters and 2.5 MB memory. The inference speed can reach 30 FPS on a single RTX 2080Ti GPU with a 1024×2048-pixel image.*

## 1. Introduction

Semantic segmentation – in which we label each pixel of an image with a corresponding class – is an important current topic in computer vision [1]. It has numerous applications in fields including autonomous driving systems and robot navigation. Most of those applications need high accuracy, however, and increasing it requires a great increase the number of parameters. Increasing the network size generally slows inference, which then may prevent the real-time operation. For example, PSPNet [2] and DeepLab [3] have good performance in semantic segmentation, but they contain tens of millions of parameters and less than 1 frame per seconds (FPS) inference speed. Obviously, these large networks are not capable of real-time (typically, 30 FPS) semantic segmentation tasks. Those real-world applications will achieve competitive performance after consideration of memory requirements and inference speed. Therefore, building an efficient segmentation network with small capacity and fast inference speed becomes an important current topic.

Existing high-speed semantic segmentation models like ENet [4] and ESPNet [5] have very high inference speeds, but at the cost of segmentation performance. Some other networks like ContextNet [6] and ICNet [7] have successfully achieved better performance, but they sacrifice the speed and model size respectively. Therefore, we seek a better balance among accuracy, inference, and model size in this study.

Much previous work [8, 9] has already demonstrated the potential of multi-scale convolution, which is able to perceive various sizes of receptive fields. This method allows the network to take advantage of multi-level feature extraction and incorporate various scales of information. In addition, dilated convolution has been shown to be a promising method to extract large-scale features while maintaining the total number of parameters [10, 11]. Both methods contain limitations, however. As a channel-wise module, the Inception module contains many parameters even though it applies factorizations. A dilated convolution with a single dilation rate can extract global information but may miss local features. In the case of the Cityscapes dataset, for example, a fixed dilation rate makes the network only extract features from large classes but prevents it from identifying small ones.

Based on the above observation and thinking, we proposed a novel tiny CNN module that combines the advantages of Inception and dilated convolution. We apply this novel module to build a shallow but effective encoder-decoder--based network to extract dense features. Our main contributions are:

- We propose an efficient module that incorporates the Inception module and dilated convolution, and they are called the Channel-wise Feature Pyramid (CFP) module. This module extracts various size feature map and contextual information jointly and significantly reduces the number of parameters and model size.
- We design the Channel-wise Feature Pyramid Network (CFPNet) based on the CFP module. It has many fewer parameters and better performance than the existing state-of-art real-time semantic segmentation network.
- We achieve very competitive results on both Cityscapes [12] and CamVid [13] benchmarks without any context module, pre-trained model, or post-processing. Using fewer parameters, the proposed CFPNet substantially outperforms existing segmentation networks. It can process high-resolution images (1024×2048) at 30 FPS on a single RTX 2080Ti GPU, yielding 70.1% class-wise and 87.4% category-wise mean intersection-over-union (mIoU) on the Cityscapes test dataset with only 0.55 million parameters.

## 2. Related Work

Recently many real-time semantic segmentation techniques that have been proposed have great potential, such as dilated convolution, factorization, and low-bit networks, to minimize network size and speed up CNNs. First, we briefly describe the techniques we used and then give a brief overview of the encoder-decoder based semantic segmentation.

**Inception module:** The naïve Inception module [8] proposed a parallel structure which contains $1 \times 1$, $3 \times 3$ and $5 \times 5$ convolution kernel to obtain multi-scale feature maps. Those large kernels, however leading to great computational cost. Therefore, a further version of the Inception module introduced factorizing convolutions to reduce the number of parameters. The factorization contains two parts: factorization into smaller convolution and asymmetric convolutions [9]. A $5 \times 5$ convolutional operator, for example is replaced by two $3 \times 3$ convolutions. And then if we factorize a standard convolution into a $3 \times 1$ convolution followed by $1 \times 3$ convolution can save 33% of parameters with the same number of filters. These two factorization methods have been applied widely and successfully shown great potential in reducing the computation of the CNN model (e.g. ResNext [14], Xception [15], and MobileNets [16]). The CFP module is inspired by these factorization approaches. The CFP module first applied factorization into a small convolution approach to greatly simplify the Inception-like module to a single CNN channel. Then, the asymmetric convolution technique used to further reduce the parameters of this channel. Factorization reduces the computation substantially, while allowing the module to learn the features from a series of sizes of the receptive field.

**Dilated convolution:** Dilated convolution [17] introduced a special form of standard $3 \times 3$ convolution by inserting gaps between pairs of convolution elements to enlarge the effective receptive field without introducing more parameters. The effective size of an $n \times n$ dilated convolution kernel with a dilation rate $r$ can be represented as: $[r(n-1)+1]^2$. The dilation rate is the number of pixel gaps between adjacent convolution elements, and only $n^2$ parameters participate in model training. Many studies have applied dilated convolution to build a spatial feature pyramid to extract a multi-scale feature representation (e.g., DeepLab series [3, 10, 11] and DenseASPP [18]). Most of the applications have shown the potential in pixel-level tasks. In our study, we also applied dilated convolution in each channel of the CFP module.

**Encoder-decoder based CNNs for semantic segmentation:** The encoder-decoder based network can be divided into two parts: encoder and decoder. Usually, the encoder is sequence of convolution and down-sampling operators used to extract high dimensional features. Then these features are decoded by up-sampling and convolution to generate segmentation masks. There are many encoder-decoder-based state-of-art implementations including U-Net [19], SegNet[20] and FCN [21] that show great potential in pixel-level segmentation tasks. Our study is related to that architecture by using the CFP module.

## 3. CFPNet

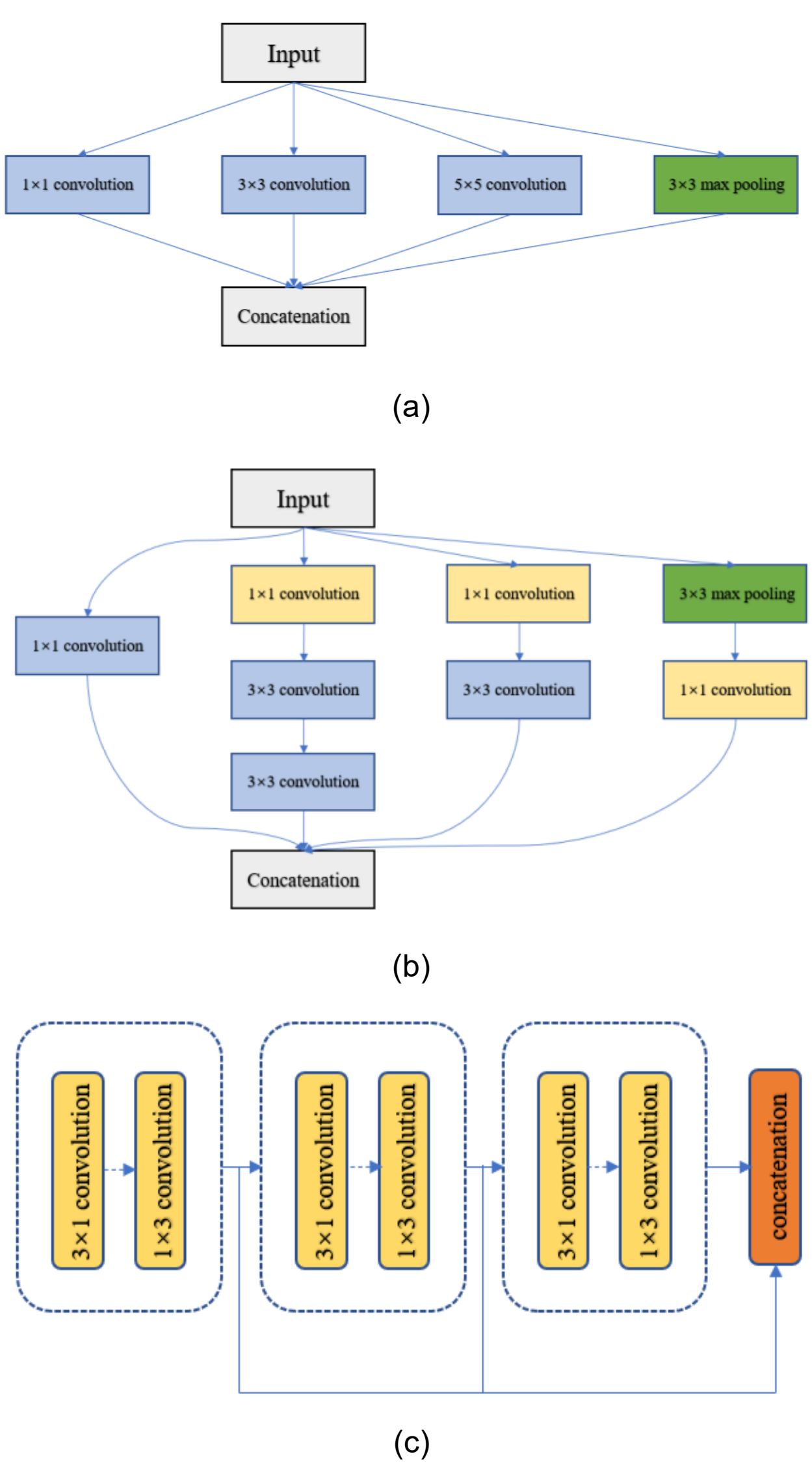

Figure 1. (a) Naïve Inception module. (b) Inception-v2. (c) Feature Pyramid channel

In this section, we first introduce the basic element of the CFP module – the Feature Pyramid channel. We then present the structure of the CFP module and architecture of CFPNet.

### 3.1 Feature Pyramid Channel

CFPNet is based on the channel-wise feature pyramid (CFP) module, a factorized form of convolution operator that decomposes a large kernel into smaller convolution as shown in Figures 1(a) and (b). The original Inception module directly uses a $5 \times 5$ size kernel, but Inception-v2 applies two $3 \times 3$ convolutional operators to replace it. Based on the ideas of multi-scale feature maps and factorizations, we want to design a module that contains up to a $7 \times 7$ kernel. Similarly to Inception-v2, we applied two and three $3 \times 3$ convolution kernels to replace $5 \times 5$ and $7 \times 7$ size kernels. Although this operation saves 28% and 45% of the parameters respectively, it still too large to achieve our real-time goals. So, we merged those convolution kernels into one channel which contains only three $3 \times 3$ kernels. Then we decomposed the standard convolution into an asymmetric form to build the Feature Pyramid (FP) channel as shown in Figure 1(c). We used a skip connection to concatenate features that are extracted from each asymmetric convolution block to create a multi-scale feature map. Compared with Inception-v2's implementation, the FP channel further saves 67% of parameters but also retains the ability to learn feature

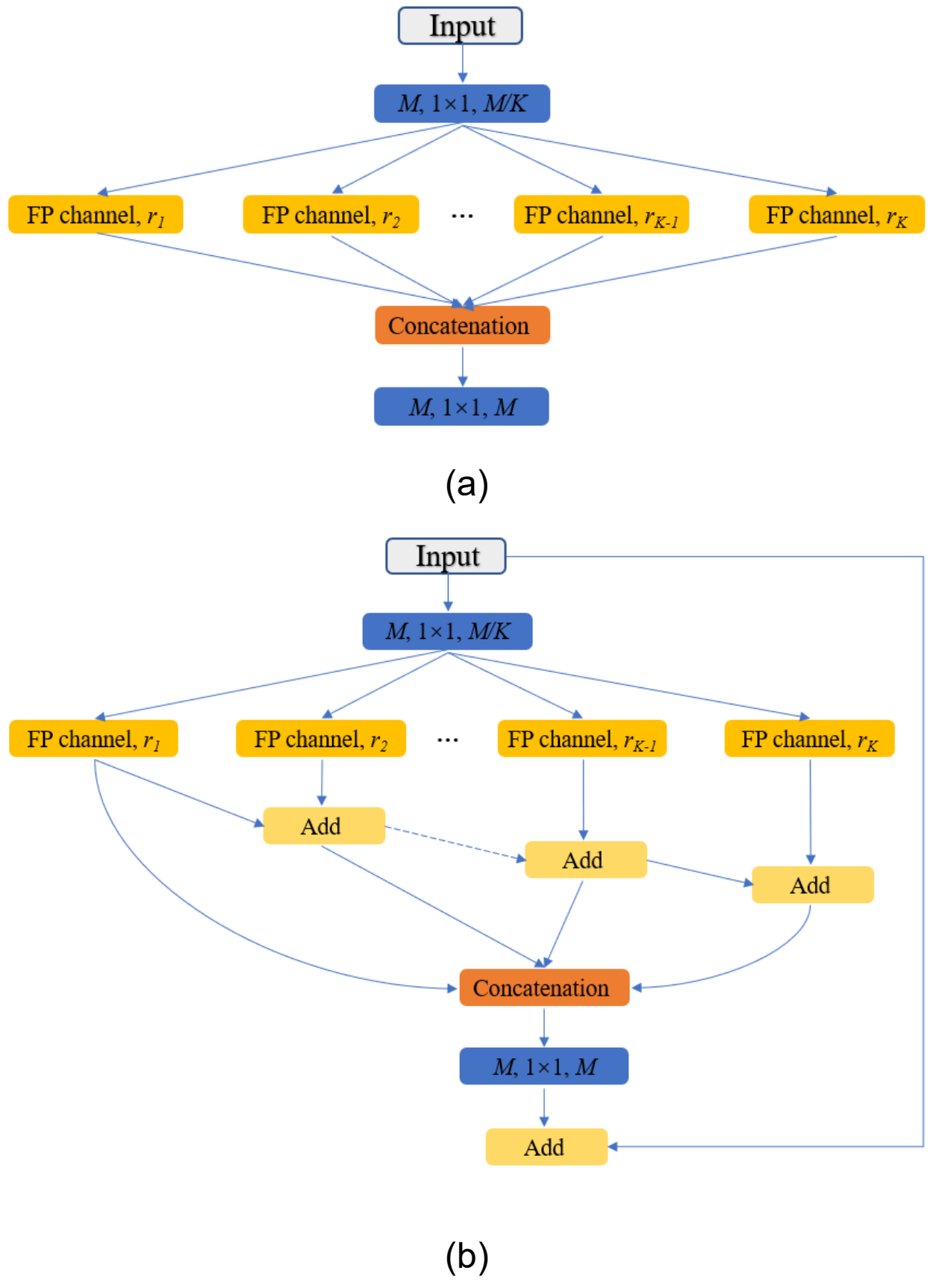

Figure 2. Structure: (a) Original CFP module. (b) CFP module

information from the same--size receptive field.

Due to the concatenating features of each asymmetric convolution block, we need to keep the same dimension of input and output by rearranging filter numbers for each asymmetric block. If the input dimension is $N$, we assigned $N/4$ for the first and second blocks which correspond to the $3 \times 3$ and $5 \times 5$ convolution, respectively. For the third block, the $7 \times 7$ kernel, we assigned $N/2$ filters to extract large weighted sizeable features.

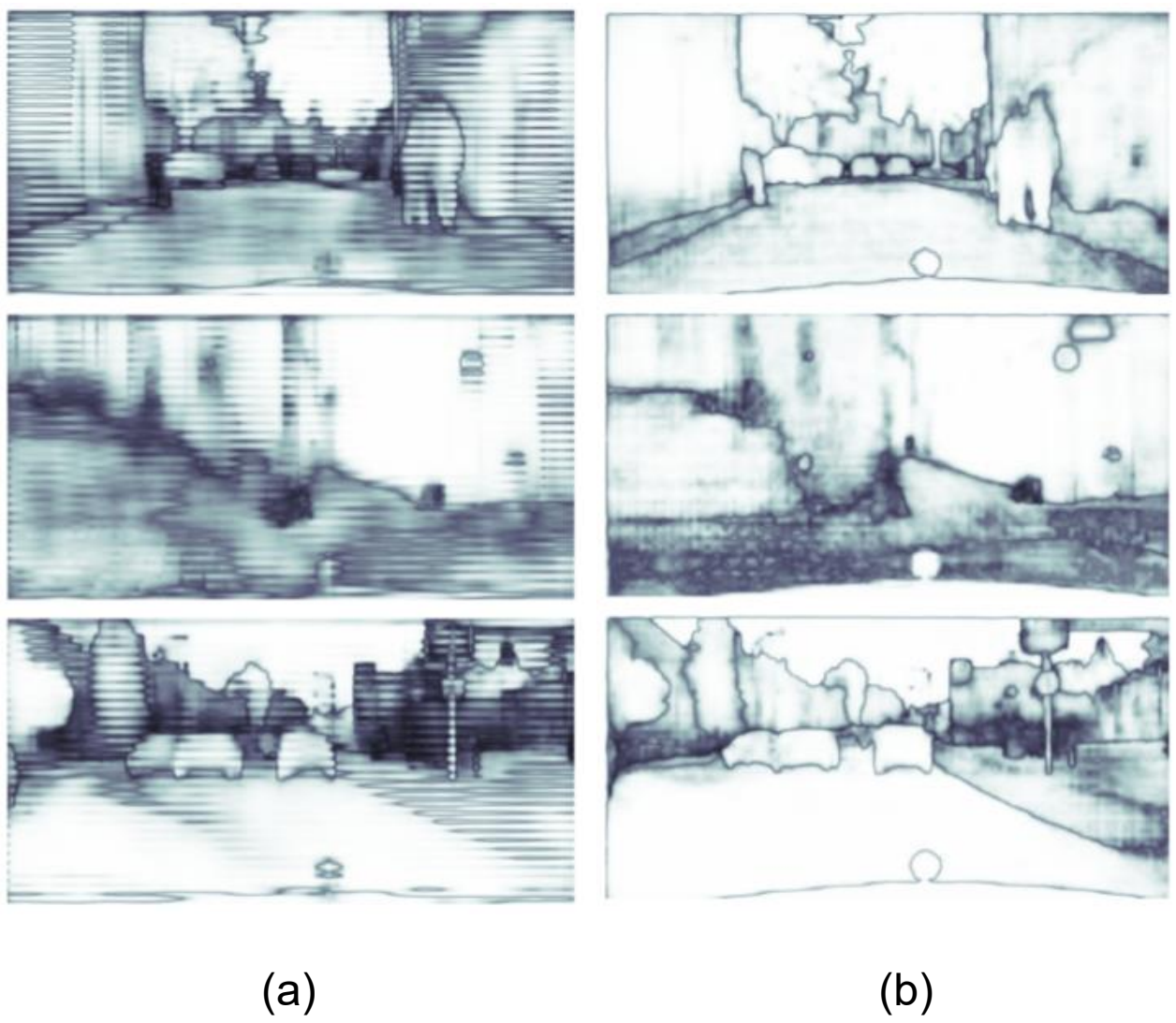

Figure 3. Hierarchical Feature Fusion [5]: (a) Without HFF. (b) With HFF

### 3.2 CFP Module

The CFP module contains $K$ FP channels with different dilation rates $\{r_1, r_2, \dots, r_K\}$. The original CFP module first applied a $1 \times 1$ convolution to reduce the input dimension from $M$ to $M/K$. Then the dimension of the first to third asymmetric block are $M/4K$, $M/4K$ and $M/2K$, respectively.

In Figure 2, we show more details about the CFP module. There, we used $1 \times 1$ convolution to project high-dimension feature maps to low-dimension. Then we set multi FP channels into a parallel structure with different dilation rates. We then concatenate all feature maps into the input's dimension and use another $1 \times 1$ convolution to activate the output. This is a basic structure of the original CFP module as shown in Figure 2(a). Asymmetric convolution increases the depth of the network, however, which makes it difficult to train. Moreover, a simple fusion method introduces some unwanted checkerboard or gridding artifacts that greatly influence the accuracy and quality of segmentation masks as shown in Figure 3(a). To solve the difficulty in training, we first used the residual connection to make a deeper network trainable and also to provide additional feature information [22]. To address effect of gridding artifacts, we applied hierarchical feature fusion (HFF) [5] to de-gridding. Starting from the second channel, we take a sum operation to combine feature maps step by step, then concatenate them to build a final hierarchical feature map. Finally, the influence of gridding artifacts is reduced as shown in Figure 3(b). The final version of the CFP module is shown in Figure 2(b).

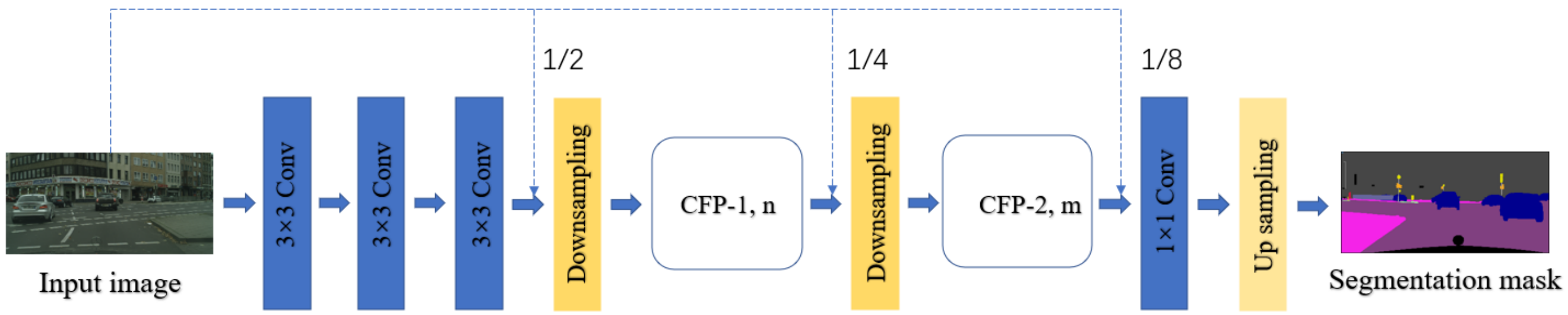

Figure 4. CFPNet: Architecture of proposed CFPNet. It contains two CFP clusters and $\boldsymbol{n}$, $\boldsymbol{m}$ represent the repeat times of these two clusters.

### 3.3 CFPNet

**Setup CFP module:** Firstly, we specify the details about the CFP module which we used to build the CFPNet. We choose the number of FP channel $K = 4$. For the input with dimension like $M = 32$, the total filter number of each channel is 8. And we set the filter numbers of the first to third asymmetric convolution block are 2, 2 and 4, respectively. Then we set different dilation rates for each FP channel. Take dilation rate equal to $r_K$, for example, we set the first and fourth channels' dilation rate is $r_1 = 1$ and $r_K$ because we want to CFP module could extract local and global features. For the second and third channels, we set dilation rates are equal to $r_2 = r_K/4$ and $r_3 = r_K/2$, respectively. So, the CFP module could also learn those medium-size features. If $r_K/4$ is less than 1, like $r_K = 2$, we directly set this channel's dilation rate is 1.

**Network architecture:** Since our goal is to build a lightweight network but have competitive performance, so we build a shallow network as shown in Figure 4. And the details of the architecture can be seen in Table 1. We first use three $3 \times 3$ convolutions as the initial feature extractor. Then we applied the same down-sampling method with ENet [4], which combines a $3 \times 3$ convolution with a stride 2 and a $2 \times 2$ max pooling. Those three times down-sampling operations produce outputs whose dimensions are $\frac{1}{8}th$ of the original size of the input. Before the first, second max pooling layers and final $1 \times 1$ convolution, we use skip connection to inject resized input images to provide additional information for segmentation network. For the CFP-1 and CFP-2 clusters, we choose the repeat times of CFP module $n = 2$ and $m = 6$ with the dilation rate $r_{K_{CFP-1}} = [2,2]$ and $r_{K_{CFP-2}} = [4,4,8,8,16,16]$. Finally, we use a $1 \times 1$ convolution to activate the final feature map and a simple decoder – bilinear interpolation to generate the final segmentation masks. All of those convolutions are followed by PReLU [23] activation function and batch normalization. Because studies have already proven that PReLU achieves better performance than ReLU in a shallow network.

| No. | Layer | Mode | Dimension |
|---|---|---|---|
| 1 | $3 \times 3$ Conv | stride 2 | 32 |
| 2 | $3 \times 3$ Conv | stride 1 | 32 |
| 3 | $3 \times 3$ Conv | stride 1 | 32 |
| 4 | Downsampling | - | 64 |
| 5-6 | 2×CFP | $r_K = 2$ | 64 |
| 7 | Downsampling | - | 128 |
| 8-9 | 2×CFP | $r_K = 4$ | 128 |
| 10-11 | 2×CFP | $r_K = 8$ | 128 |
| 12-13 | 2×CFP | $r_K = 16$ | 128 |
| 14 | $1 \times 1$ Conv | stride 1 | 19 |
| 15 | Bilinear interpolation | ×8 | 19 |

Table 1. Architecture details of CFPNet

## 4. Experiments

In this section, we test our proposed neural network on two challenging datasets: Cityscapes and CamVid which are widely used in semantic segmentation. We introduce these two datasets first and implementation protocol. Then we test our network by several experiments on CamVid to determine those parameters such as repeat times, dilation rates and the number of channels etc. Then, we test its performance on the Cityscapes test dataset and report official results. Finally, we make a comparison with some other state-of-art networks.

### 4.1 Datasets

**Cityscapes.** The Cityscapes focus on semantic understanding of urban scenes. It contains 5000 fine annotation and 20000 coarser annotation images. And the dataset was captured from 50 different cities in different seasons and weathers. For the fine annotation set, it contains 2975 training, 500 validation and 1525 testing images. The resolution of original images is $1024 \times 2048$. The whole datasets contain 19 classes that belong to 7 categories (e.g. car, truck and bus belong to the vehicle category). We finally evaluated our test performance through the Cityscapes online server.

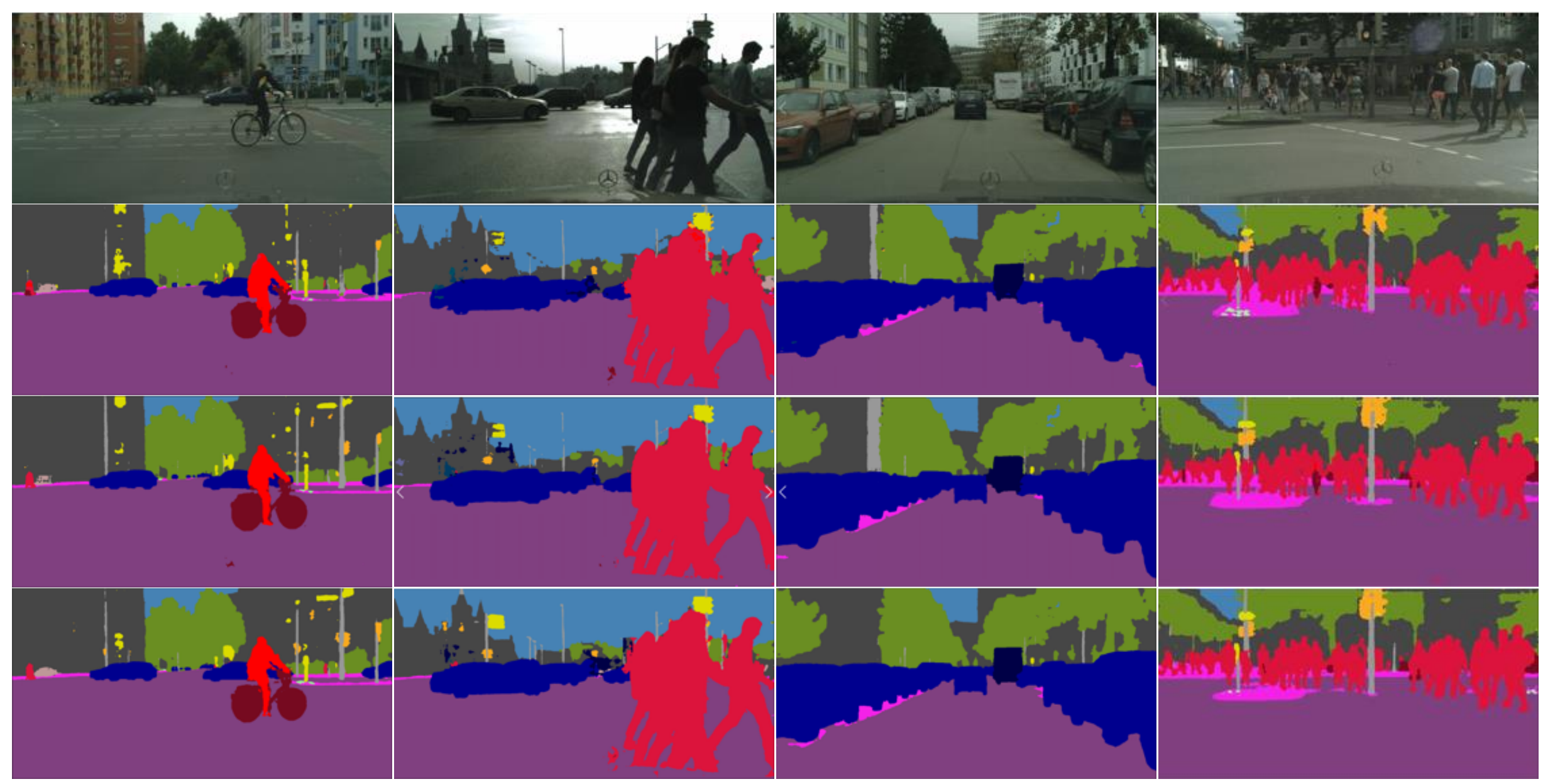

Figure 5. Segmentation examples. From top to bottom is original image, CFPNet-V1, CFPNet-V2 and CFPNet-V3

| Net | Roa | Sid | Bui | Wal | Fen | Pol | TLi | TSi | Veg | Ter | Sky | Per | Rid | Car | Tru | Bus | Tra | Mot | Bic | mIoU |
|---|---|---|---|---|---|---|---|---|---|---|---|---|---|---|---|---|---|---|---|---|
| V1 | 96.7 | 76.3 | 88.0 | 36.0 | 46.9 | 50.2 | 43.9 | 58.5 | 89.8 | 55.1 | 91.2 | 69.0 | 41.5 | 89.7 | 39.5 | 54.4 | 23.7 | 26.6 | 63.2 | 60.4 |
| V2 | 97.4 | 79.8 | 89.9 | 47.4 | 50.7 | 55.3 | 53.9 | 67.2 | 90.9 | 59.2 | 93.4 | 74.5 | 48.6 | 92.3 | 49.9 | 63.1 | 31.1 | 49.3 | 69.9 | 66.5 |
| V3 | **97.8** | **81.4** | **90.5** | 46.4 | 50.6 | **56.4** | **61.5** | **67.7** | **92.1** | **68.9** | **94.3** | **80.4** | **60.7** | **93.9** | **51.4** | **68.0** | **50.8** | **51.2** | 67.7 | **70.1** |

Table 4. Class-wise comparison on Cityscapes test set. V1, V2 and V3 represents CFPNet-V1, 2, 3.

**CamVid.** The CamVid dataset is also an urban scene dataset that is used for autonomous driving applications. It includes 701 images with resolution $720 \times 960$, 367 for training, 101 for validation and 233 for testing. It contains 11 classes, and we resize those images into $360 \times 480$ before training.

## 4.2 Implementation protocol

CFPNet is trained by using PyTorch [24] with CUDA 9.0 and cuDNN V7. We evaluated the inference time on a single RTX 2080Ti GPU by calculating the average runtime of 100 frames.

For network training, we choose ADAM [25] as the optimizer with momentum 0.9 and weight decay $4.5e^{-4}$ in training. Here we applied the "poly" learning rate policy [26] and the initial learning rate is set with power 0.9. We choose a different batch size for two datasets, 8 for Cityscapes and 16 for CamVid. Also, we choose cross-entropy as a loss function to train network. Because we do not use any pre-train methods, so we set maximum training epochs is 1000. To increase the diversity of training, we apply data augmentation approaches. It contains random horizontal flip, mean subtraction, and random scale on input images. The scale rates are {0.5, 0.75, 1.0, 1.25, 1.5, 1.75}. After scaling, we randomly crop images into a fixed input size.

| Network | Parameter | Size | mIoU |
|---|---|---|---|
| CFPNet-V1 | 0.31 M | 1.34 MB | 61.6% |
| CFPNet-V2 | 0.37 M | 1.6 MB | 62.9% |
| CFPNet-V3 | 0.55 M | 2.5 MB | **64.3%** |

Table 2. Evaluation of CFPNet on CamVid.

## 4.3 Experiments of CFPNet Architecture

In this section, we proposed various CFPNet which have different channel numbers, repeat times and dilation rates, and evaluated their performance CamVid test dataset. Details of the CFPNet are shown in Table 2.

**CFPNet-V1.** CFPNet-V1 is the shallowest version

because its repeat times $n$ and $m$ are 1 and 2. For the first CFP cluster, we set dilation rates $r_{K_{CFP-1}} = [4]$ and $r_{K_{CFP-2}} = [8,16]$.

**CFPNet-V2.** Compared with the CFPNet-V1, we hope the network can extract more local features. So, we change the repeat time from $\{n, m\} = \{1, 2\}$ to $\{n, m\} = \{1, 3\}$, and corresponding dilation rates are modified to $r_{K_{CFP-1}} = [2]$ and $r_{K_{CFP-2}} = [4,8,16]$.

**CFPNet-V3.** To improve the performance of network and control model size. We double the repeat times compared with CFPNet-V2. And the dilation rates in each cluster are modified to $r_{K_{CFP-1}} = [2,2]$ and $r_{K_{CFP-2}} = [4,4,8,8,16,16]$.

In addition, we also test these three versions on Cityscapes test set. The evaluation results are shown in Table 3. We could find both on CamVid and Cityscapes dataset CFPNet-V3 get the best performance.

Figure 5 shows some segmentation results on Cityscapes test set. From top to bottom: Original image, prediction of CFPNet-V1, CFPNet-V2 and CFPNet-V3. And we also print the result of each class as shown in Table 4.

| Network | Parameter | Size | mIoU |
|---|---|---|---|
| CFPNet-V1 | 0.31 M | 1.34 MB | 60.4% |
| CFPNet-V2 | 0.37 M | 1.6 MB | 66.5% |
| CFPNet-V3 | 0.55 M | 2.5 MB | **70.1%** |

Table 3. Evaluation of CFPNet on Cityscapes.

## 4.4 Results and Comparison

In this subsection, we compare our method with other networks. Firstly, we report the results on Cityscapes and CamVid test dataset, then we analyze the accuracy, parameters, model size and inference speed.

**Accuracy and model size.** We first analyze the relationship between segmentation accuracy and model size. We compare CFPNet with some state-of-art networks on Cityscapes test set. Their relationships between size and accuracy are shown in Figure 6 and Table 5.

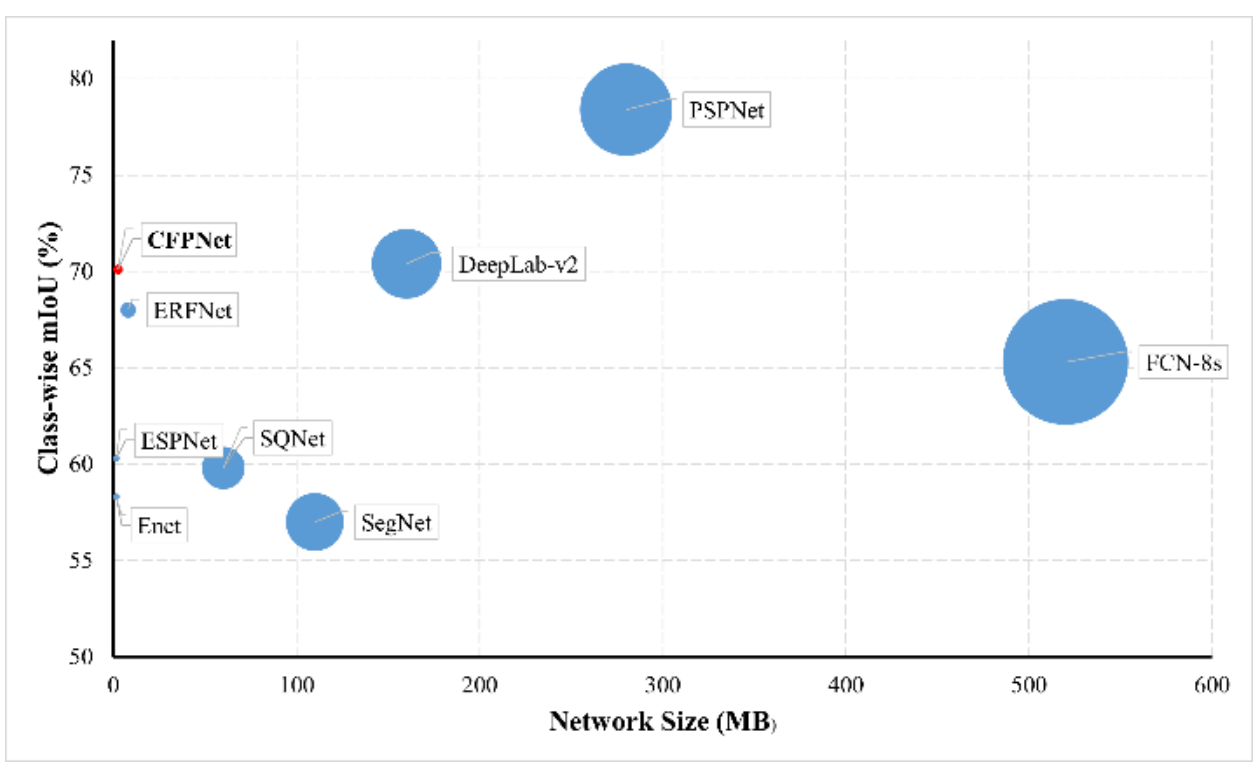

(a)

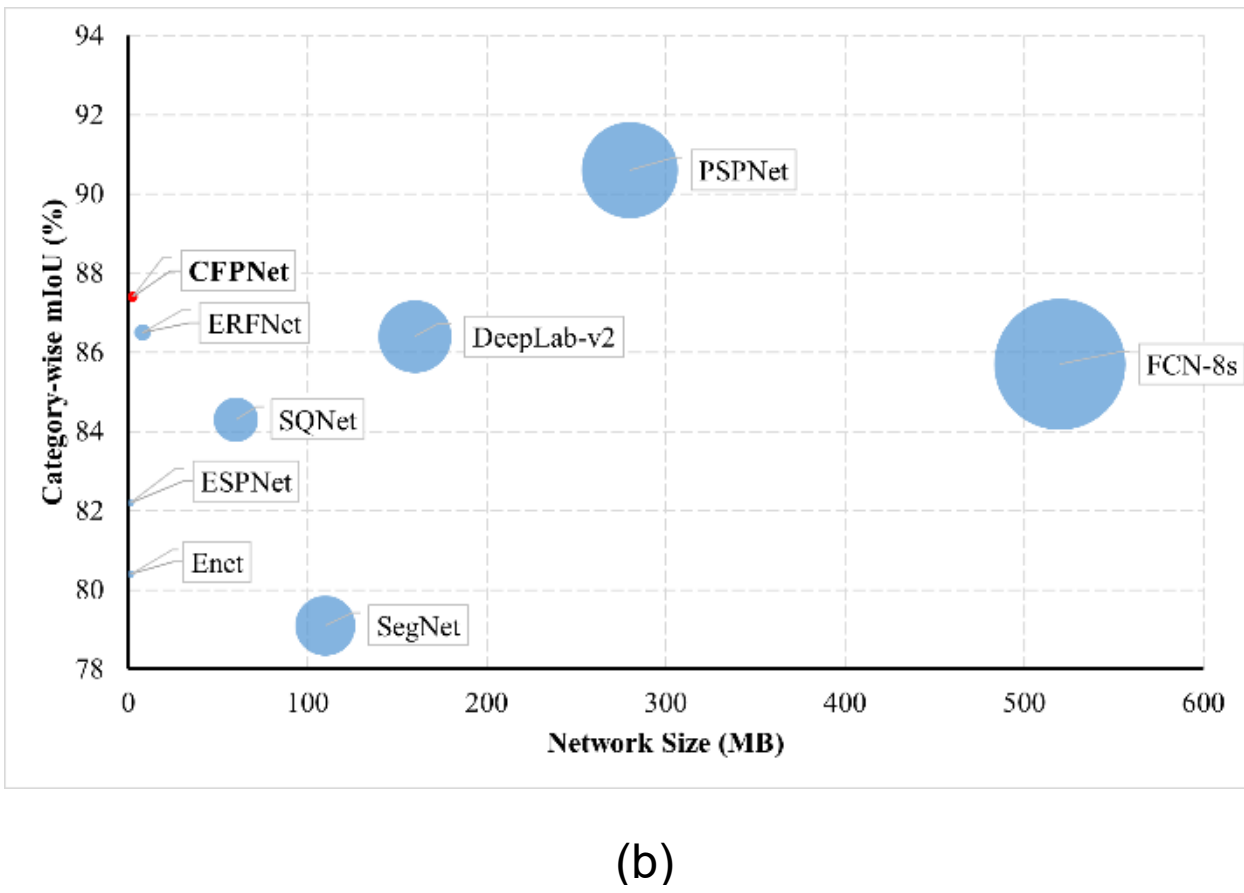

(b)

Figure 6. (a) Network size and class-wise mIoU accuracy. (b) Network size vs category-wise mIoU accuracy.

| | mIoU (%) | |
|---|---|---|
| **Network** | **Class** | **Category** |
| ENet [4] | 58.3 | 80.4 |
| ESPNet [5] | 60.3 | 82.2 |
| ERFNet [27] | 68.0 | 86.5 |
| SQNet [28] | 59.8 | 84.3 |
| SegNet [20] | 57.0 | 79.1 |
| FCN-8s [21] | 65.3 | 85.7 |
| DeepLab-v2 [3] | 70.4 | 86.4 |
| PSPNet [2] | 78.4 | 86.4 |
| **CFPNet** | 70.1 | 87.4 |

Table 5. Results on Cityscapes test set

In Figure 6, the area size of the circle represents the model size (e.g. smaller circle represents a smaller model size). It is obviously that although the size of the CFPNet is tiny, it has very competitive mIoU accuracy both for class-wise and category-wise as shown in Table 5. We can find that CFPNet-V3 much more sensitive and accurate for small and low-frequency class like the traffic light.

**Accuracy and parameters.** We continue to make a comparison between accuracy and parameters. The test results on Cityscapes dataset can be seen in Figure 7.

The area of the circle represents the number of parameters. We can find in the networks whose parameters less than 1 million, CFPNet has a very high accuracy which achieves 70.1%. Its performance is much better than the ENet, ESPNet and CGNet [29] in the same parameter level. Moreover, CFPNet has fewer parameters than DABNet [30] and LEDNet [31] which have similar performance. Compared with those networks whose parameters greater than 5 million like BiSeNet [32], ICNet and DeepLab-v2, CFPNet even get a better segmentation accuracy.

**Accuracy and inference speed.** Since different networks use different input size and GPU. So we report both input size and GPU types in Table 6. The GPU computational ability is: TitanX Maxwell < Titan X

| Network | Pretrain | InputSize | mIoU (%) | FPS | Parameters | GPU |
|---|---|---|---|---|---|---|
| DeepLab-v2 [3] | ImageNet | 512 × 1024 | 70.4 | $< 1$ | 44 | TitanX |
| PSPNet [2] | ImageNet | 713 × 713 | 78.4 | $< 1$ | 65.7 | TitanX |
| SegNet [20] | ImageNet | 360 × 640 | 56.1 | 14.6 | 29.5 | TitanX |
| ENet [4] | None | 512 × 1024 | 58.3 | 76.9 | 0.4 | TitanX |
| SQ [28] | ImageNet | 1024 × 2048 | 59.8 | 16.7 | - | TitanX |
| ESPNet [5] | None | 512 × 1024 | 60.3 | 112 | 0.4 | TitanX-P |
| ContextNet [6] | None | 1024 × 2048 | 66.1 | 18.3 | 0.85 | TitanX |
| ERFNet [27] | None | 512 × 1024 | 68.0 | 41.7 | 2.1 | TitanX |
| BiSeNet [32] | ImageNet | 768 × 1536 | 68.4 | 105.8 | 5.8 | TitanXp |
| ICNet [7] | ImageNet | 1024 × 2048 | 69.5 | 30.3 | 7.8 | TitanX |
| CGNet [29] | None | 360 × 640 | 64.8 | 50 | 0.5 | 2×V100 |
| LEDNet [31] | None | 512 × 1024 | 70.6 | 71 | 0.94 | 1080Ti |
| DABNet [30] | None | 1024 × 2048 | 70.1 | 27.7 | 0.76 | 1080Ti |
| **CFPNet** | None | 1024 × 2048 | **70.1** | **30** | **0.55** | 2080Ti |

Table 6. Evaluation results on the Cityscapes test set. TitanX represents the TitanX Maxwell, TitanX-P represents the TitanX Pascal and 2080Ti represents RTX 2080Ti

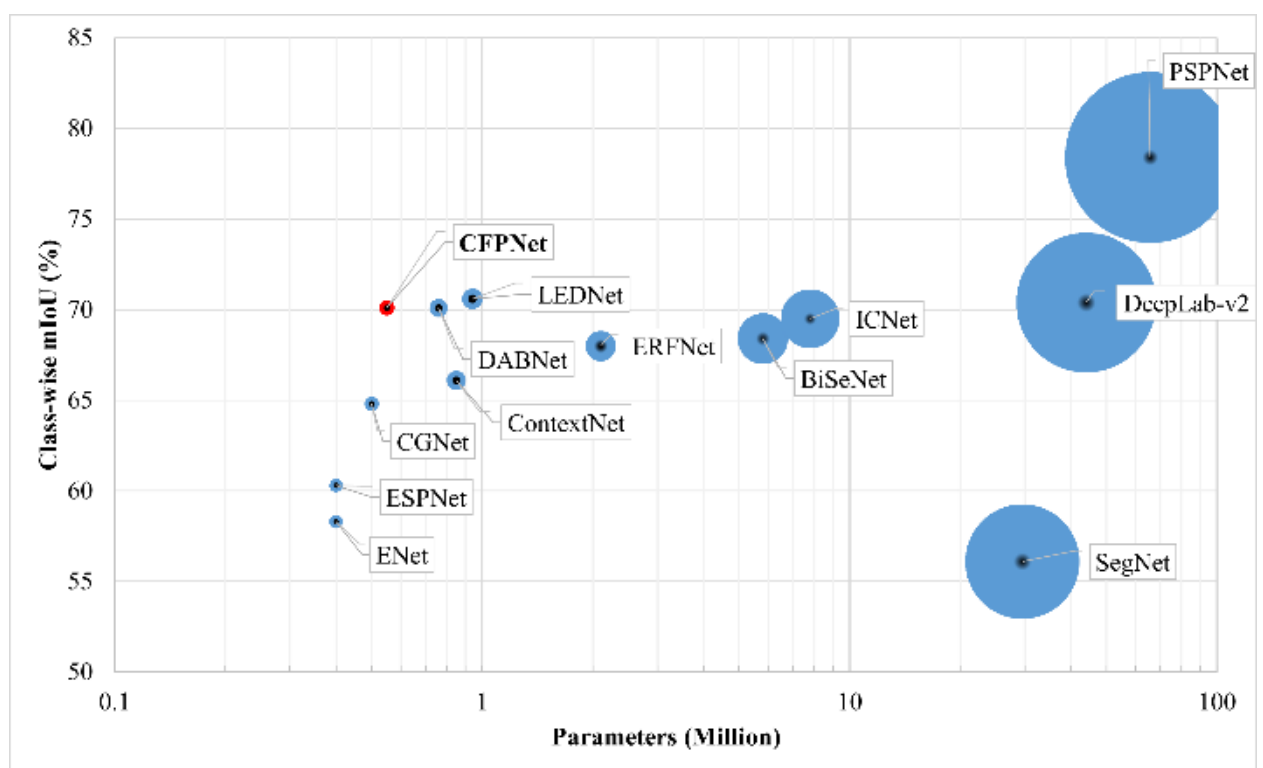

Figure 7. Parameters and class-wise mIoU

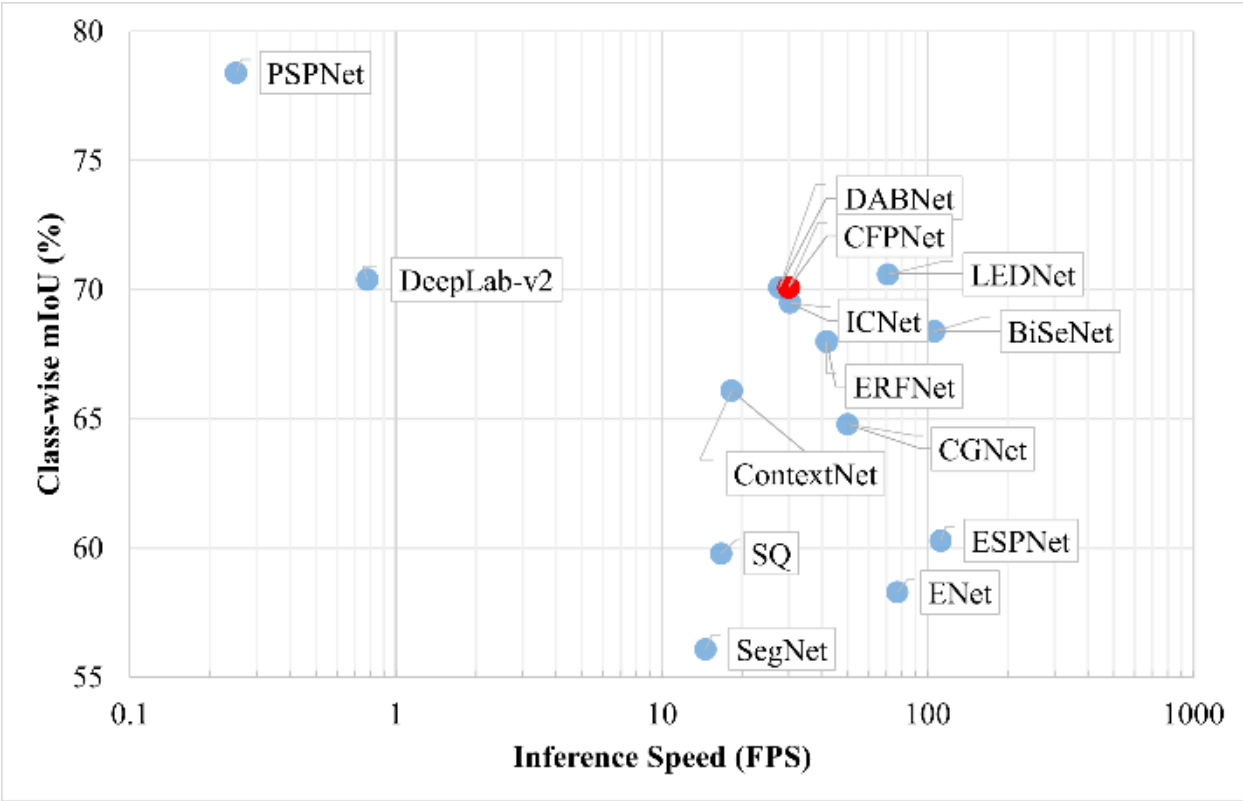

Figure 8. Inference speed and class-wise mIoU.

Pascal ≈ GTX 1080Ti < TitanXp < RTX 2080Ti < V100. Although the input size and GPU devices are different, we still plot Figure 8 to help compare those results from Table 6.

In terms of inference speed, we can find CFPNet has a similar processing speed with DABNet and ICNet which both have 1024 × 2048 input size. However, the parameter of CFPNet is only 7.1% of ICNet's. Although CFPNet and DABNet have similar mIoU accuracy and inference speed, CFPNet saves 28.6% parameters than DABNet to achieve this goal. Compared with some other high-speed networks like ENet and ESPNet, the segmentation performance of CFPNet exceeds them by over 10%. For those classical state-of-art networks like SegNet, DeepLab-v2, and SQ etc. CFPNet has much better performance on both accuracy and inference speed.

We also test the performance on CamVid test dataset and compare it with some other state-of-arts. As we report in Table 7, CFPNet also achieves outstanding performance with a small size. Take ENet and ESPNet for example, they have the least amount of parameter, but it significantly influences their performance which are lowest in Table 7. Compare with other high-performance methods, CFPNet merely has less than 10% parameters but achieves competitive accuracy.

| Network | mIoU (%) | Parameters (M) |
|---|---|---|
| ENet [4] | 51.3 | 0.36 |
| SegNet [20] | 55.6 | 29.5 |
| FCN-8s [21] | 57.0 | 134.5 |
| Dilation8 [33] | 65.3 | 140.8 |
| BiSeNet [32] | 65.6 | 5.8 |
| ESPNet [5] | 55.6 | 0.36 |
| CFPNet | 64.7 | 0.55 |

Table 7. Performance on CamVid test set

## 5. Conclusion

In this paper, we proposed a tiny real-time semantic segmentation network, CFPNet which is based on the Feature Pyramid channel. We analyze the efficiency of CFPNet from its model size, parameters, inference speed and mIoU accuracy. Both the analysis and experimental results on Cityscapes and CamVid dataset show the potential of CFPNet as a tiny, fast and efficient network. We also test it performance in the medical image area, and it shows a very good performance both on accuracy and parameters. All of those shows CFPNet is a novel efficient network for semantic segmentation.